\documentclass[10pt,letterpaper]{article}

\usepackage[english]{babel}
\usepackage[utf8]{inputenc}
\usepackage[T1]{fontenc}

\usepackage{booktabs}
\usepackage{array}
\usepackage{longtable}
\usepackage{multirow}
\newcolumntype{L}[1]{>{\raggedright\arraybackslash}p{#1}}

\usepackage[letterpaper,left=1.5in,right=1.5in,top=1in,bottom=1in]{geometry}

\usepackage{amsmath}
\usepackage{amssymb}
\usepackage{newtxtext,newtxmath}
\usepackage{microtype}

\makeatletter
\newcommand{\plainpercent}{{\usefont{T1}{lmr}{\f@series}{n}\char37}}
\makeatother
\renewcommand{\%}{\plainpercent}

\usepackage[compact]{titlesec}
\titleformat{\section}{\large\bfseries}{\thesection}{0.6em}{}
\titleformat{\subsection}{\normalsize\bfseries}{\thesubsection}{0.6em}{}
\titlespacing{\section}{0pt}{1.2\baselineskip}{0.5\baselineskip}

\usepackage{graphicx}
\usepackage{xurl}

\usepackage[colorlinks=true, allcolors=blue]{hyperref}

\hypersetup{
  pdftitle={A Contract-Grade Verifier for LLM-Generated GPU Kernels, and a Native
            Blackwell Backward for the Gated-Linear-Recurrence Family},
  pdfauthor={Rishi Shah and Rishav Shrestha},
  pdfsubject={Preprint},
  pdfkeywords={GPU kernels; kernel verification; LLM code generation; numerical
               correctness; benchmark auditing; KernelBench; CUDA; Triton; Blackwell;
               tcgen05; state space models; gated linear attention; Mamba},
}

\usepackage{float}
\usepackage[font=small,labelfont=bf,labelsep=colon,skip=6pt]{caption}

\title{A Contract-Grade Verifier for LLM-Generated GPU Kernels,
and a Native Blackwell Backward for the Gated-Linear-Recurrence Family}

\author{%
  \begin{tabular}{c@{\hspace{0.9in}}c}
    \textbf{Rishi Shah}            & \textbf{Rishav Shrestha}          \\
    Machine Learning Engineer      & Chief Technical Officer           \\
    E3A Healthcare                 & E3A Healthcare                    \\
    \texttt{rishishah994@gmail.com} & \texttt{rishav@e3ahealth.com}    \\
  \end{tabular}
}

\makeatletter
\renewcommand{\maketitle}{%
  \null\vskip 6pt
  \begin{center}
    {\rule{\linewidth}{1.4pt}}\vskip 8pt
    {\LARGE\bfseries \@title \par}\vskip 8pt
    {\rule{\linewidth}{0.5pt}}\vskip 22pt
    {\normalsize \@author \par}\vskip 16pt
  \end{center}
  \vskip 20pt
}
\makeatother

\newenvironment{abstractblock}{%
  \begin{center}{\bfseries\normalsize Abstract}\end{center}
  \vspace{-0.4em}
  \begin{list}{}{\setlength{\leftmargin}{0.4in}\setlength{\rightmargin}{0.4in}}
  \item[]%
}{%
  \end{list}\vspace{0.8em}
}

\begin{document}
\maketitle

\begin{abstractblock}
Systems that generate GPU kernels with language models report high correctness rates. Those
rates come from a single loose test: run the kernel on a few random inputs at one fixed shape
and accept it if the output is close to a reference. A kernel can pass that test and still be
silently wrong. It can return an ordinary number where the true answer is a NaN or an infinity,
produce a different result on each run, break the moment the shape changes, or accumulate in
\texttt{fp16} where the reference keeps an \texttt{fp32} total. We build the instrument that
checks correctness properly: a \emph{contract-grade verifier} of twelve adversarial gates, each
a property a correct kernel must satisfy, several of them \emph{tolerance-free} so that no choice
of threshold can explain a failure away. We aim it in two directions. Aimed outward, the verifier
audits \textbf{2{,}638} machine-generated kernels that a public system's own harness had already
accepted as correct. It finds \textbf{39.5\%} broken beyond any tolerance argument, and
\textbf{62.1\%} carrying at least one violation. The field's standard test accepts
\textbf{1{,}487} kernels the verifier rejects, against 14 in the reverse direction. Four
independent defenses hold the finding up: a \textbf{7/7} positive control, a
threshold-calibration sweep, \textbf{98.5\%} agreement with the reference benchmark's own
correctness code, and a stratified hand-audit. Aimed inward, the same battery judges a kernel of
our own: the first native Blackwell \texttt{tcgen05} training backward for the
gated-linear-recurrence (GDN) family, including the reverse-state stage open implementations run
on a fallback. That kernel's correctness is established independently of the verifier, against a
double-precision oracle, and five family members train through it, so the battery is aimed at a
subject already known to be correct. The correctness signal behind reported progress in kernel
generation is far weaker than the numbers suggest, and a set of tolerance-free contracts would
close most of the gap.
\end{abstractblock}

\section{Introduction}
\label{sec:intro}

Generating GPU kernels with large language models has become an active research program, and
its benchmarks report that a large fraction of outputs are correct and often faster than a
reference~\cite{ouyang2025kernelbench, li2025tritonbench, drkernel2026, lange2026robustkbench,
li2026cudal1}. Those reports are only as trustworthy as the correctness signal beneath them, and
that signal is almost universally a single test: draw a few random inputs at one fixed shape and
accept the kernel if its output is close, in the \texttt{allclose} sense, to a reference. A kernel
can pass this test while swallowing a non-finite, returning a different answer on each run,
breaking at the next shape, or accumulating a reduction in \texttt{fp16} against an \texttt{fp32}
reference. If the acceptance signal is this weak, an unknown share of the reported wins is
illusory.

This paper is about that gap and the instrument that measures it. We build a \emph{contract-grade
verifier} that operationalizes the Kernel Contracts taxonomy~\cite{veit2026contracts}: twelve adversarial gates, each a property a correct kernel must satisfy, graded against
a slow high-precision reference. Some gates need no tolerance at all and compare by exact mask or
byte equality; the rest use tolerances derived from a floating-point error model. The verifier is the spine of the paper, and we aim it in two directions.

Aimed outward, the verifier audits a public corpus of machine-generated kernels that the source
system's own harness had already accepted as correct, and measures how far those acceptances
overstate correctness: 39.5\% are broken in a way no tolerance can excuse. The obvious objection
is that our checker is simply stricter than everyone else's, so foreign kernels fail by
construction. Section~\ref{sec:defenses} answers it four independent ways: a positive control on a
kernel we can vouch for independently, a threshold-calibration sweep, 98.5\% agreement with
KernelBench's own correctness code~\cite{ouyang2025kernelbench}, and a stratified hand-audit of
disputed cases.

Aimed inward, the same verifier judges a kernel of our own: a hand-written native Blackwell
\texttt{tcgen05} training backward for the gated-linear-recurrence (GDN) family, whose
reverse-state stage open implementations still run on a fallback. Its correctness is established
against an independent double-precision oracle, which makes the verifier's verdict on it a
positive control: the same battery that rejects the foreign corpus clears a kernel already known
to be correct. The two uses meet in the data. The failure modes that most often sink the foreign
kernels are the ones our own kernels were at risk of, and one of those gates surfaced a real
defect in our own code. A checker that rejects our failed ideas and clears a kernel we can vouch
for independently gives weight to its acceptances and diagnostic value to its rejections.

\paragraph{Contributions.}
\begin{itemize}
  \item A \textbf{twelve-gate contract-grade verifier} (Section~\ref{sec:verifier}) operationalizing
  the Kernel Contracts taxonomy~\cite{veit2026contracts} at scale, with derived thresholds where a
  threshold is needed at all, and a red team of nineteen deliberately-broken kernels the battery
  must reject while clearing the correct reference.
  \item An \textbf{at-scale rigor-gap audit} of 2{,}638 accepted kernels: a tolerance-free floor of
  \textbf{39.5\%}, \textbf{62.1\%} carrying a contract violation, four independent defenses, and a
  differential in which the standard test certifies \textbf{1{,}487} broken kernels against 14 in
  the reverse direction.
  \item The \textbf{first native Blackwell \texttt{tcgen05}/tensor-memory training backward for
  the GDN family} (Section~\ref{sec:native}), reverse-state stage included: verified to
  $3.3\times10^{-3}$ against an \texttt{fp64} oracle, bit-for-bit deterministic, training five
  family members through one backward, and clear of the tensor-memory constraint behind \#904.
  \item \textbf{Supporting results} (Appendix~\ref{sec:support}): six contract-verified Mamba-3
  kernels that bypass \#904 on Blackwell, a 1.1B-parameter model trained end-to-end through the C5
  fused-block path, a from-scratch GRPO system that autotunes them, and the speed baselines.
\end{itemize}

Section~\ref{sec:discussion} stress-tests these results and states their limitations. Every
quantitative claim is backed by a committed artifact under a fixed software environment, and a
number resting on an unreleased intermediate is marked at the point of use.

\paragraph{Code and data availability.} Everything this paper is built on is public at
\url{https://github.com/RishiShah99/lethe} under the MIT license: the verifier and its twelve
gates, the nineteen deliberately-broken kernels of Section~\ref{sec:verifier}, the audit harness,
the six Mamba-3 Triton kernels, the native Blackwell backward, the GRPO trainer, and the JSON
evidence behind every number reported here. The audit ships as its own per-row record, one line
per audited kernel carrying the corpus identifier, operator class, run status, recorded speedup,
and the verdict and reason of every gate that fired. The three selection filters, the 2{,}638
denominator, the per-gate counts, and the tolerance-free floor are all recomputable from that one
file, whose identifier column joins directly against the corpus, so any individual row cited here
can be pulled and re-run. Two things sit outside it: the external check's own per-row verdicts,
which the differential of Section~\ref{sec:diff} and the 98.5\% agreement of
Section~\ref{sec:defenses} are joined against, ship as aggregated cell counts and a disagreement
list, and the aggregation predicates are specified as prose in Appendix~\ref{app:selection}. The
audited corpus is the public MIT-licensed \texttt{hkust-nlp/drkernel-coldstart-8k}
release~\cite{drkernel2026}, pinned at revision \texttt{cba0ef0}; the audit ran on a B200 under
torch 2.12 / triton 3.7, and the harness is public, so the run reproduces on any B200 under that
pin.

\section{Background and Related Work}
\label{sec:related}

\paragraph{Generating GPU kernels, and how their correctness is judged.} A line of work trains
or prompts language models to emit GPU kernels that replace a reference operator, and grades
them on suites such as KernelBench~\cite{ouyang2025kernelbench} and
TritonBench~\cite{li2025tritonbench}, or on system-specific corpora such as Dr.\ Kernel /
KernelGYM~\cite{drkernel2026} and the Sakana AI CUDA Engineer
archive~\cite{sakana2025archive}. Across these, correctness is decided by an approximate
equality check (\texttt{torch.allclose} at \texttt{atol}${=}$\texttt{rtol}${=}10^{-2}$ in the
common KernelBench setting) over a handful of random inputs at a single fixed shape. This is
inexpensive and reproducible, but it probes a narrow slice of a kernel's behavior.

\paragraph{The correctness illusion, and how our work differs.} The observation that loose
checks accept broken kernels is not new. Sarkar~\cite{sarkar2026illusion} argues the point on
a hand-built set of twenty-four kernels, and reports no rate on any accepted corpus. A companion
paper~\cite{sarkar2026calibration} targets the threshold itself, calibrating a per-(operator,
dtype) \texttt{atol} empirically as the 95th percentile of the error observed over passing runs of
the \emph{correct} kernel, scaled by a 1.5 safety factor. It is the strongest existing form of the
"just tighten the tolerance" answer, and its own numbers mark that answer's limits: detection
rises from 1{,}805 of 2{,}467 (73.2\%) to 2{,}034 of 2{,}467 (82.4\%), false positives on correct
controls rise from 0 of 1{,}882 to 20 of 1{,}882, and the evaluation runs on author-seeded
variants. Three differences follow. Their threshold is \emph{fitted}, so judging an operator first
requires a corpus of known-correct runs of it; ours is derived analytically from how rounding
error accumulates over a reduction, and is available for an operator seen once. Their gain is
bought with false positives on correct kernels, a cost a tolerance-free gate cannot incur, having
no threshold to miscalibrate. And a calibrated \texttt{atol}, however tight, still compares
magnitudes, so a swallowed NaN, a nondeterministic output, and an aliased buffer all lie outside
what any tolerance can express. The two lines are complementary: calibration sharpens the
thresholded gates, which is why the tolerance-free ones carry the load-bearing claim here.

The nearest prior \emph{instrument} is \texttt{robust-kbench}~\cite{lange2026robustkbench}, which
reaches the same diagnosis we do, that existing kernel benchmarks carry exploitable loopholes and
too narrow a set of testing conditions, and answers it with a harder benchmark and an agentic
pipeline whose correctness filtering is performed by \emph{language-model verifiers}. The
diagnosis is shared; the instrument is what separates us. A verifier that is itself a model
returns a judgement, so its false-accept rate is a property of that model on that kernel and must
be established separately. Ours is twelve fixed gates whose verdicts are exact mask, byte, or
derived-bound comparisons against the reference operator, with no model in the loop and, on the
tolerance-free channels, no threshold either. That is what lets us state a rate on a corpus
someone else has already accepted, and lets a reader check that rate without adopting our model.

A separate strand documents \emph{performance} loopholes, kernels that game the timing harness
by reusing an output buffer, returning a lazily-evaluated tensor, or exploiting a fixed input,
in the Sakana AI CUDA Engineer~\cite{sakana2025archive} and CUDA-L1~\cite{li2026cudal1} systems,
with a companion line of work hardening harnesses against such exploits~\cite{zhong2026hardening};
that axis is orthogonal to the numerical silent-wrongness we measure. The Kernel Contracts
taxonomy~\cite{veit2026contracts} enumerates correctness classes for GPU kernels; the taxonomy is
theirs, and ours is the first runnable, at-scale operationalization of it that we are aware of.
Three things separate this paper from the literature above: a \emph{tolerance-free} correctness
measure, a quantified audit on thousands of \emph{accepted, real} kernels, and a certified native
artifact the same verifier passes.

\paragraph{The gated-linear-recurrence family.} Sub-quadratic sequence models replace
attention's quadratic cost with a fixed-size recurrent state. Mamba~\cite{gu2023mamba} and its
structured-state-space-duality successor Mamba-2/SSD~\cite{dao2024mamba2}, gated linear
attention (GLA)~\cite{yang2024gla}, DeltaNet~\cite{yang2024deltanet} and its gated
variant~\cite{yang2025gateddeltanet}, and Mamba-3~\cite{mamba3} all instantiate one
\emph{gated linear recurrence}. Section~\ref{sec:native} makes this precise: each is a fixed
setting of that recurrence's gates, so a single backward differentiating the general form yields
the training gradients for all of them. That backward is dominated by two hard stages, a
reverse-time inter-chunk state scan and a WY / triangular-inverse vector-Jacobian product, and the
reference open-source implementation of both is the Triton \texttt{flash-linear-attention}
(\texttt{fla}) library~\cite{fla2024}, which is our speed baseline.

\paragraph{Blackwell \texttt{tcgen05}, tensor memory, and \#904.} NVIDIA's Blackwell
generation (the B200, architecture \texttt{sm\_100}) adds a fifth-generation tensor core,
\texttt{tcgen05}, whose matrix-multiply operands reside in a new scarce on-chip space, Tensor
Memory (TMEM), governed by a hard 512-column budget per warpgroup. A multi-GEMM kernel must
manage TMEM allocation and release lifecycles by hand; a lifecycle error yields illegal machine
code or a deadlock. This constraint is the crux of the open issue
\texttt{state-spaces/mamba\#904}~\cite{mamba904}: the official Mamba-3 backward uses the
tensor-core matmul (\texttt{tl.dot}), and at the normal performance setting a compiler pass
requests 544 TMEM columns against the hardware's 512, so the kernel fails to compile at those
settings and falls back to a path the issue reports as $38.7\times$ slower on GB200. That issue
attributes the slowdown to a \texttt{ptxas} \texttt{C7907} error eliminating autotuner
configurations. On our stack the tensor-memory half reproduces and the C7907 half does not
(Appendix~\ref{sec:904}). A merged Triton change (PR \#9093) addresses the tensor-memory half, and
it has not shipped in any Triton 3.7.x release, so a pinned pre-\#9093 stack still reproduces it. Both headline
contributions engage this constraint: the verifier's RES-02 gate detects it, and the native
backward gets the lifecycle right where the official kernel does not.

\section{A Contract-Grade Verifier and the Rigor-Gap Audit}
\label{sec:verifier}

\subsection{The twelve gates}
\label{sec:gates}

A kernel that passes a single random-input, fixed-shape check can still be wrong on extreme
inputs, wrong at a different shape, nondeterministic, secretly low-precision, or wrong about
non-finite values. The verifier therefore asks twelve questions, each an adversarial contract
graded against a slow reference (Table~\ref{tab:gates}). For our own operators the reference is a
plain high-precision loop, \emph{defined} to be ground truth, which refuses to run in reduced
precision so that it can never be misused as a candidate. In the outward audit the reference is
the corpus row's own \texttt{fp32} PyTorch module, held to the same role by the mixed-precision
convention of Section~\ref{sec:audit}: for a half-dtype probe it consumes inputs and parameters
round-tripped once through that dtype and then computes in \texttt{fp32}, so the answer key is
always a full-precision computation.

Two design choices make the battery rigorous. First, every tolerance is derived. ORD-01's bound
follows how floating-point error accumulates over a reduction,
$\text{atol}\approx 4\,\varepsilon\sqrt{N}\cdot\text{scale}$, so a correct kernel passes with wide
margin and a wrong one misses by orders of magnitude, and every knob sits at the loosest point
that still sides with the candidate. Second, the input draw is seeded, so a verdict depends on the
kernel and not on luck. The two entry points seed differently. Called directly, as the red team
and our own kernels' acceptance gates call it, the battery re-seeds from a fixed value per gate,
mixing in the gate's name so that no two gates see the same stream. The corpus audit fixes one
seed per worker process and lets the draws run in gate order, which is what the released verdicts
were frozen against; a verdict there is a deterministic function of the (reference, candidate)
pair and the gate order, and not of the gate in isolation.

On a forward-only inference corpus, CMP-02 (gradient) has no autograd graph to check and RES-02
(resource metadata) has no compile-time metadata to read, so ten of the twelve gates carry the
audit. Of those ten, seven are \emph{load-bearing}: CMP-01/03, ORD-01/02, EXC-01/02, and PRC-01.
Each of the other three is set aside for a stated reason. ORD-03 is the deliberately-relaxed
backstop whose coverage is carried by CMP-01 and ORD-01; PRC-02 is graded by a separate
accumulator mechanism outside the band-gate scheme; and RES-01 is low-signal at 1.8\%. The two
idle gates do run against our own kernels: the committed test suite exercises the full twelve on
them, supplying compiled-kernel resource metadata for RES-02 and per-gradient views so CMP-02 has
an autograd graph to differentiate. The positive control of Section~\ref{sec:defenses} is judged
by the same ten-gate audit subset as the corpus, so control and corpus pass through identical
machinery.

The repository ships nineteen deliberately-broken kernels, each embodying one trick: returning its
input, caching an answer, accumulating in \texttt{fp16}, behaving nondeterministically, flipping
an infinity to a finite value, working at only one shape, or silently moving data to another
device. A test asserts that the verifier rejects every one \emph{and} clears the correct
reference. That two-sided check is what establishes the referee before it is ever aimed at foreign
code.

\begin{table}[H]
\centering\small
\caption{\label{tab:gates}The twelve contract gates, operationalizing the Kernel Contracts
taxonomy~\cite{veit2026contracts}. Gates in the \emph{tolerance-free} group (EXC-01, EXC-02,
ORD-02, plus aliasing and crash detection) compare by exact mask or byte equality and admit no
tolerance argument; the rest use scale-aware derived tolerances. The reported floor drops EXC-02
from that group and adds a non-artifact failure to run, which is no gate at all;
Appendix~\ref{app:floor} gives the channel list and the effect of including EXC-02.}
\begin{tabular}{@{}l >{\raggedright\arraybackslash}p{11.4cm}@{}}
\toprule
Gate & Property checked \\
\midrule
CMP-01 & correct on many random \emph{and} adversarial inputs (zeros, $10^{6}$, $10^{-6}$, denormals, long $L$) \\
CMP-02 & the \emph{gradients} are correct, not only the outputs (autograd versus finite difference) \\
CMP-03 & correct across shapes (batch, length, width), not only the one tested \\
ORD-01 & reordered summation stays within a derived $\propto\!\sqrt{N}$ rounding bound \\
ORD-02 & byte-for-byte identical across five repeats, and does not alias a shared buffer \\
ORD-03 & correct on an input constructed to expose a bad summation order \\
PRC-01 & correct in \texttt{fp32}, \texttt{fp16}, and \texttt{bf16} \\
PRC-02 & fed \texttt{fp16}, still keeps an internal \texttt{fp32} running total \\
EXC-01 & fed an input with NaN and signed infinities scattered at known positions, the output's non-finite positions and signs match the reference exactly \\
EXC-02 & flush-to-zero handling of subnormals matches the reference \\
RES-01 & output lives on the same device as the input \\
RES-02 & the compiled kernel fits real hardware limits (registers, shared memory, TMEM budget) \\
\bottomrule
\end{tabular}
\end{table}

\subsection{The audit}
\label{sec:audit}

We ran the verifier over Dr.\ Kernel / KernelGYM~\cite{drkernel2026}
(\texttt{hkust-nlp/drkernel-coldstart-8k}, MIT license), a public corpus of 8{,}920
supervised-fine-tuning trajectories each ending in a Triton kernel that replaces a PyTorch
module in the KernelBench \texttt{Model}/\texttt{ModelNew} convention. It was the richest
auditable release among the systems we surveyed. Its SSM-adjacent operator classes (matmul,
attention, softmax, scan, norm, conv, reduction) yield \textbf{3{,}134} kernels, each run on a
B200 (torch 2.12 / triton 3.7) in its own sandboxed subprocess. Three filters then reduce that
set to the headline denominator, in this order. First, \textbf{2{,}970} carry a positive recorded
speedup ($\texttt{final\_speedup}>0$), the corpus's own record that the kernel cleared its
pipeline and was timed. The release ships no separate correctness column, so this is the strongest
acceptance predicate the artifact supports; Section~\ref{sec:defenses} reports one sampled row
where a speedup was recorded for a candidate whose output our harness could not read at all, which
bounds how far that predicate can be read as a correctness certificate. Second, \textbf{2{,}919}
of those expose a floating-point tensor input for the gates to perturb; the 51 without one are
marked not-applicable and leave the denominator. Third, \textbf{2{,}638} remain after excluding
toolchain and compile artifacts. This accepted-only set is the headline denominator.
Appendix~\ref{app:selection} gives the operator classifier and the artifact predicate exactly, so
each of the three counts is reconstructible.

Fairness is enforced in code by five rules, pinned by nineteen unit tests. Inputs the
\emph{reference} itself cannot run are marked not-applicable and never counted as candidate
failures; tolerances sit at the loosest defensible point; positions where candidate and reference
agree on a NaN or an infinity go uncharged, since a non-finite admits no tolerance comparison,
while a \emph{swallowed} non-finite is still caught by EXC-01; a candidate is
charged only for its own declared compute precision; and the mixed-precision and shape conventions
are fixed in advance. Every ambiguous case moves toward the candidate. One precondition cuts the
other way. EXC-01 scatters non-finites into the input, so it probes behavior a kernel's author may
never have intended to support. The reference defines the contract: every kernel in this
population is advertised as a drop-in replacement for one specific PyTorch module, so if that
module propagates a non-finite and the replacement hands back an ordinary number, the operator's
behavior has changed, whatever case can be made that the change is an improvement.

\begin{figure}[H]
\centering
\includegraphics[width=0.86\textwidth]{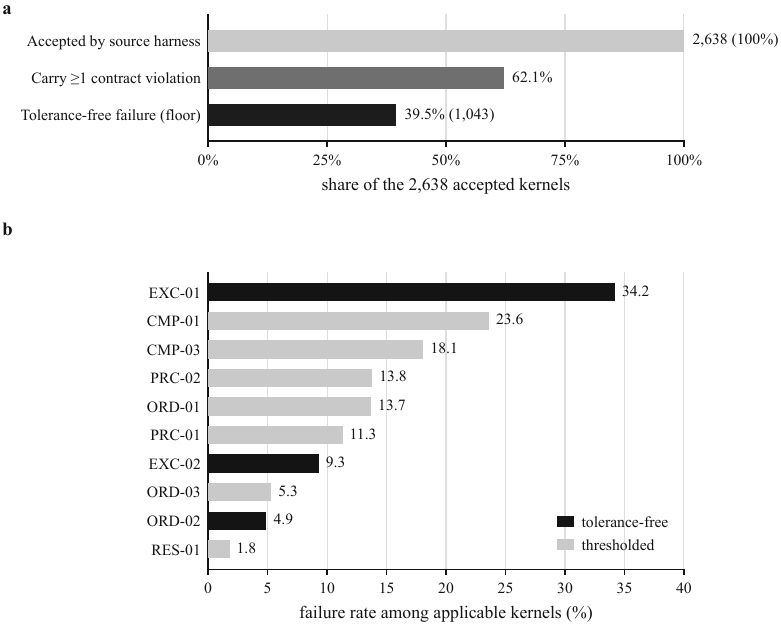}
\caption{\label{fig:audit}The rigor-gap finding. \emph{Top:} of the 2{,}638 kernels a public
system's own harness accepted as correct, 62.1\% carry at least one contract violation and
39.5\% (1{,}043) fail a tolerance-free gate, the floor no tolerance argument can reach.
\emph{Bottom:} per-gate failure rate among the kernels where each gate applies (tolerance-free
gates in solid black), over per-gate applicability denominators, so these rates are not additive
to the corpus-level floor above. Exact counts are in Appendix~\ref{app:numbers}.}
\end{figure}

\paragraph{Headline.} Among the 2{,}638 accepted kernels, \textbf{62.1\%} carry at least one
contract violation, and \textbf{39.5\%} (1{,}043 of 2{,}638) fail a \emph{tolerance-free} gate,
broken in a way no tolerance argument can excuse. Figure~\ref{fig:audit} presents the finding as
an acceptance funnel together with the per-gate failure rates (exact counts in
Appendix~\ref{app:numbers}). The modal defect is a kernel silently replacing a NaN or infinity
with an ordinary number (EXC-01), the failure mode that converts a training-time error into silent
data corruption. That channel carries the floor: 868 of the 1{,}043 floor members fail EXC-01, and
the floor without it is 229 kernels (8.7\%), which makes the floor a non-finite-propagation
result. The 62.1\% carries the breadth, and it survives the same ablation: EXC-01 accounts for 868
of the 1{,}639 kernels with a violation, and deleting the gate entirely leaves \textbf{42.9\%}
(1{,}132 of 2{,}638), still above the tolerance-free floor this paper leads with. No other gate
moves the figure by more than five points (Table~\ref{tab:pergate}). One bookkeeping detail
follows from how the two figures are computed. The 62.1\% runs over gate verdicts and run
failures, while output aliasing is tracked outside that union despite being a floor channel, so
nine aliasing-only kernels sit inside the 39.5\% and outside the 62.1\%. Folding aliasing into
ORD-02, where Table~\ref{tab:gates} already places it, gives 1{,}648 of 2{,}638 (62.5\%) and makes
the floor a strict subset. We report the smaller number in both directions.

Both figures survive aggressive re-slicing. Adding CMP-03 shape rejection to the floor raises it
to 41.1\%, a further 40 kernels; the floor admits only outright rejection of a shape, since a
value mismatch at a new shape still carries a tolerance argument. The floor's one run-failure
channel, the 62 kernels that fail to execute for a reason the artifact filter does not catch, is
not carrying it either: some of those 62 are toolchain-shaped in a way the four-prefix predicate
misses, and deleting the channel outright leaves 981 of 2{,}576, or 38.1\%. The 62.1\% violation
rate holds at 61.9\% after stripping the entire matmul class, the worst case for the
\texttt{TF32} tolerance argument, and at 54.7\% after stripping every gate to which a tolerance
argument could be attached at all (ORD-01, PRC-01, PRC-02, EXC-02).

\subsection{Four independent defenses of the finding}
\label{sec:defenses}

The objection raised in Section~\ref{sec:intro} is that the checker is simply stricter than
everyone else's, so foreign kernels fail by construction. Four answers follow, each falsifiable on
its own.

\paragraph{(1) Positive control: 7/7.} Our own six Mamba-3 Triton kernels (Appendix~\ref{sec:c6})
and the native GDN backward (Section~\ref{sec:native}) run through the \emph{same} battery in
the \emph{same} \texttt{Model}/\texttt{ModelNew} convention. All seven pass every applicable gate
and are clean on every tolerance-free channel, the exact classes that sink the foreign corpus.
One of the seven is independent by construction: the tolerance model was calibrated against the
six Triton kernels, and the native GDN backward played no part in that calibration. The
tolerance-free channels carry no threshold to calibrate at all, which makes a clean result there
calibration-independent for all seven.

The control caught a gap in our own code. C5 (\texttt{fused\_block\_forward}) inferred its channel
dimension from the input tensor without checking it against the convolution-weight channels, a
missing input-validation guard. Commit \texttt{7f46226} adds the guard in three files: the two
public ops, and the \emph{reference} as well, since a depthwise convolution with mismatched
channels is outside this block's math on either side. Guarding the reference changes C5's CMP-03
verdict from a failure to not-applicable, under the reference-applicability rule applied
corpus-wide, and that is the norm here: six of the seven control kernels resolve CMP-03 as
not-applicable for the same reason. The control then reads 7/7 over uneven applicability, nine
gates applying to all seven kernels and passing on all seven (CMP-01, ORD-01/02/03, PRC-01/02,
EXC-01/02, RES-01) while CMP-03 applies to one. What carries the load is that the gate flagged the
author's own kernel and the fix changed shipped code on both sides of the comparison.

\paragraph{(2) Threshold calibration.} This is the graded version of the positive control. For each
band-gate on a known-correct operator, we inject a perturbation of known magnitude and locate
the exact pass-to-fail trip point, bracketed between the noise floor (the error a correct
kernel emits) and the real-error onset (the error a wrong kernel emits). Every band-gate shows
clean separation (Figure~\ref{fig:calib}, with margins tabulated in
Appendix~\ref{app:numbers}). One margin is thin: ORD-03 sits $1.2\times$ below its real-error
onset, being the deliberately-relaxed gate, and its coverage is carried by CMP-01 and ORD-01.

\begin{figure}[H]
\centering
\includegraphics[width=0.82\textwidth]{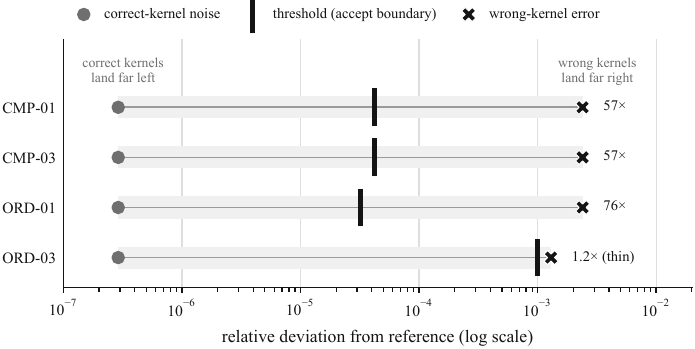}
\caption{\label{fig:calib}Threshold calibration. For each band-gate, the chosen threshold
(bar) sits inside the safe margin between the noise a correct kernel emits (circle, left) and
the error a wrong one emits (cross, right), on a log scale of relative deviation from the
reference. Every threshold clears a correct kernel and catches a wrong one. ORD-03's upper margin
is thin ($1.2\times$). PRC-02 is graded by a different mechanism (an \texttt{fp16}-carry
accumulator crosses its threshold for $N{\ge}2048$) and is omitted here.}
\end{figure}

\paragraph{(3) The benchmark's own harness agrees (98.5\%).} To settle whether our replica of the
standard check is faithful to it, we ran KernelBench's \emph{own} correctness
code~\cite{ouyang2025kernelbench} at the pinned commit \texttt{48642c5c} on the accepted pairs
and joined \emph{their} verdict against our reimplementation. Agreement is \textbf{98.5\% over
1{,}030 pairs} (844 pass/pass, 171 fail/fail, 15 disagreements), covering 39\% of the accepted
set; the sharded run was stopped once agreement had stabilized, at 98.9\% over 458 pairs and
98.5\% over 1{,}030, its expected value being near 100\% and its residual uncertainty at that
sample size unable to reach the headline. All 15 disagreements resolve. Seven are our replica
being \emph{stricter}, which cannot inflate the finding, and all seven are in the \texttt{norm}
class. The other 8 are rows our replica accepts and their code rejects: 2 where their harness
raised before any comparison happened (a bool-tensor subtraction and a Triton compile error) and 6
input-draw-dependent value mismatches, whose maximum differences run from $0.14$ to $5.10$ across
five operator classes. The bound is 8 rows of 1{,}030 in one direction against 7 in the other, a
net of one row inside the sample. The two patches this required were scoped to \emph{loading}
(device-index resolution, and importing from a temporary file so Triton traces the source), so the
correctness logic under test is byte-for-byte theirs. Their code ran on 1{,}030 of the 2{,}638
while the differential of Section~\ref{sec:diff} uses our replica over all of them, so what this
check establishes is agreement at 98.5\% wherever both ran.

\paragraph{(4) Stratified hand-audit.} We hand-traced 31 disputed cases (accepted by the
benchmark, rejected by us), stratified across every channel, and classified each from the gate
semantics and the failure signature: \textbf{16 are broken outright}, \textbf{8 are real but
tolerance-dependent} (the class a loose \texttt{allclose} hides, and the reason the differential
cell exists), and \textbf{7 fall out of scope} and leave the strict floor. Of the 16, \textbf{11}
fail on a formally tolerance-free channel and 2 more mint a NaN from a finite input, which no
tolerance excuses even though a thresholded gate is what caught it; the remaining 3 hold on the
evidence but rest on a threshold, and stay out of the floor's language. The sample was drawn
before the applicability and artifact filters of Section~\ref{sec:audit}, over a frame of 1{,}533
against the 1{,}487 the cell holds once those filters apply, a 46-row gap, which is why one of the
31 is a row the audit itself goes on to exclude.

Three identifiers, all resolvable in the released per-row file, show the range. Id 394 is a matmul
that hands back the same output storage across calls: its recorded $4.04\times$ speedup and its
aliasing are the same fact, so the number that made it attractive is the defect. Id 28 is an
RMSNorm that returns a NaN when fed a finite $10^{6}$ input, a non-finite minted out of ordinary
numbers. Both were accepted by the standard check at both its paper-era and its hardened
tolerance. Id 518 is one of the seven discards: its operator takes no floating-point tensor input
for the gates to perturb, so the verifier marks it not-applicable and it leaves the denominator,
and yet the external timing loop still recorded a $36\times$ ``speedup'' for it. That is evidence
about timing-first acceptance, it says nothing about silent wrongness, and it is counted in
neither figure.

\subsection{The differential}
\label{sec:diff}

\begin{figure}[H]
\centering
\includegraphics[width=0.66\textwidth]{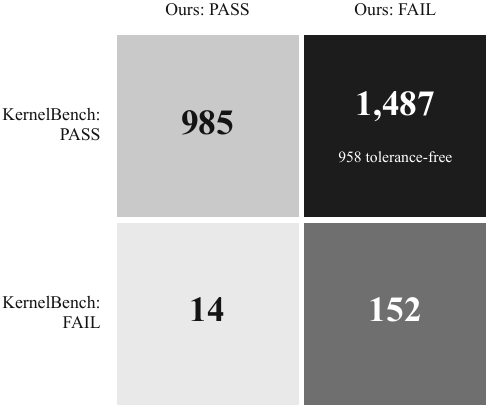}
\caption{\label{fig:diff}The differential, a $2\times2$ contingency table with each cell
shaded by its count. KernelBench's own paper-era check accepts 93.7\% of these kernels;
the load-bearing cell (accepted by the benchmark, rejected by us) holds 1{,}487 kernels, 958 of
them on a tolerance-free gate, while only 14 go the other way. The near-unidirectional
disagreement is the signature of a systematic blind spot in the acceptance signal.}
\end{figure}

Running the same accepted kernels through KernelBench's standard paper-era check
(\texttt{allclose} \texttt{atol}${=}$\texttt{rtol}${=}10^{-2}$, five random-input trials, fixed
shapes)~\cite{ouyang2025kernelbench} produces the $2\times2$ contingency table of
Figure~\ref{fig:diff}. The benchmark accepts \textbf{93.7\%} (2{,}472 of 2{,}638) of these
kernels. The load-bearing cell, external PASS and our FAIL, is \textbf{1{,}487} (56.4\% of the
accepted set), of which 958 fail on a tolerance-free gate. Both checks draw random inputs and 126
of these kernels are measurably nondeterministic, so the cell moves slightly on a re-run, and we
release two full runs of the same 2{,}638-row join: the second reads 1{,}482 with 954
tolerance-free, and 15 in the reverse cell against 14. The cell is stable to about five rows, well
inside the distance to any conclusion drawn from it. The reverse cell, external FAIL and our PASS,
holds \textbf{14} (0.5\%), so the two checks disagree almost entirely in one direction, a pattern
a uniformly stricter \texttt{allclose} would not produce. Under KernelBench's \emph{hardened}
per-dtype variant (\texttt{fp32} tolerance $10^{-4}$) the accept rate falls to 84.6\%, and
\textbf{1{,}263} kernels still pass their check and fail ours, so the finding survives the
tightened threshold. The standard test certifies nearly 1{,}500 broken kernels as correct.

\subsection{Robustness: a second stack and a second corpus}
\label{sec:robust}

The finding reproduces on a different software stack. A 300-kernel sample re-audited on torch
2.11.0+cu128, against the audit's 2.12 and on the same B200 silicon, returns \textbf{68.6\%} over
the 245 rows that survive the same accepted-only and artifact filters, above the 62.1\% headline.
Composition does not explain the gap: the high-rate reduction class is \emph{under}-weighted in
the re-audit (26.5\% of rows against 28.7\%), and standardising the re-audit to the main corpus's
class mix moves it further up, to 68.9\%. The per-class rates are higher on the older stack in six
of the seven classes, reduction at 84.6\% against 71.8\% and softmax at 59.4\% against 47.4\%, so
the newer toolchain understates the finding if anything. The pre-registered kill criterion was a
re-audit rate below 5\%, and the observed 68.6\% clears it by more than thirteen times.

A second corpus of \emph{native CUDA} kernels, the Sakana AI CUDA Engineer
archive~\cite{sakana2025archive}, shows a related but \emph{weaker} pattern. The selection rule is
the archive's Level-1 split, restricted to rows the system's own \texttt{Correct} boolean marks
correct and that expose the KernelBench \texttt{Model}/\texttt{get\_inputs} convention our harness
needs; the first 214 such rows were audited, one timed out in the sandbox, leaving 213. Its raw
100\% fail rate is precision-dominated, the kernels being \texttt{fp32}-only by design, so the
reportable figure is the environment-robust residual: \textbf{20.2\%} (43 of 213), driven by
CMP-03 shape rigidity (37), EXC-02 subnormal handling (9), ORD-02 nondeterminism (4), and CMP-01
value (1). Those per-gate counts total 51 across 43 distinct kernels, since 6 kernels fail more
than one non-precision gate and one fails four. Silent wrongness is absent here (CMP-01 1/213,
median maximum difference $0.001$). The same rigor gap recurs in native CUDA along a
complementary and weaker axis, precision and shape rigidity.

\section{A Native \texttt{tcgen05} Backward for the GDN Family}
\label{sec:native}

The second contribution is a piece of systems software that the verifier and a double-precision
oracle jointly certify: a hand-written native Blackwell \texttt{tcgen05} tensor-memory training
backward for the gated-linear-recurrence family.

\subsection{One recurrence, five models}
\label{sec:family}

A Mamba-style selective scan keeps a vector state $h$ and updates
$h_t = \bar a_t \odot h_{t-1} + \bar b_t \odot u_t$, $y_t = C_t^{\top} h_t + D\odot u_t$, where
the per-step decay $\bar a_t$ and write $\bar b_t$ are recomputed from the input at every step.
That input dependence, known as selectivity, is what makes the model expressive and what forbids
precomputing the scan as one matrix multiply. The models we target use a matrix state $S$ of
shape $d_k\times d_v$ and the stronger gated DeltaNet (GDN) update. Writing $q_t,k_t$ for
queries and keys, $v_t$ for values, and three gates $g_t$ (per-channel log-decay), $b_t$
(erase), and $w_t$ (write),
\begin{equation}
S_t = \big(I - k_t (b_t \odot k_t)^{\top}\big)\,\mathrm{Diag}(e^{g_t})\,S_{t-1}
      + k_t (w_t \odot v_t)^{\top},
      \qquad o_t = S_t^{\top} q_t .
\label{eq:gdn}
\end{equation}
Equation~\eqref{eq:gdn} combines channel-wise decay, the delta rule (the model subtracts what
$S$ already predicts for a key before writing, so it stores the correction rather than the raw
value), a rank-one write, and a read with $q_t$. The delta rule couples every step to the entire
state, and the backward must unwind that coupling in reverse.

Equation~\eqref{eq:gdn} is a superset. Fixing its gates recovers named models
(Table~\ref{tab:family}), so one backward differentiating \eqref{eq:gdn} once produces the
training gradients for all five. Each member is re-derived independently from its own definition
and checked against \emph{its own} reference, with no knob-setting on the GDN path standing in for
that check. Two reductions are exact enough to serve as built-in tests: $b{=}w{=}\beta$ reduces
\eqref{eq:gdn} to the KDA equation, pinned in \texttt{fp64} at $10^{-12}$ because the two paths
contract in different orders, and $b{=}0$ makes an entire triangular-solve stage vanish
($M{=}0\Rightarrow T{=}I$), which is pinned bitwise.

\begin{table}[H]
\centering\small
\caption{\label{tab:family}One recurrence, five models. Each named model is a gate setting of the
general recurrence, so a single hand-written backward trains them all, and each is verified
against its own independent reference.}
\begin{tabular}{@{}l>{\raggedright\arraybackslash}p{6.2cm}l@{}}
\toprule
Model & Recurrence $S_t = \dots$ & gate setting \\
\midrule
LA (linear attention)        & $S_{t-1} + k_t v_t^{\top}$ & $g{=}b{=}0,\ w{=}1$ \\
GLA (gated linear attention)  & $\mathrm{Diag}(e^{g_t})\,S_{t-1} + k_t v_t^{\top}$ & $b{=}0,\ w{=}1$ \\
SSD / Mamba-2 (scalar decay)  & $e^{g_t} S_{t-1} + k_t v_t^{\top}$ & $b{=}0,\ w{=}1$, scalar $g$ \\
KDA (Kimi Delta Attention)    & $(I-\beta_t k_t k_t^{\top})\,\mathrm{Diag}(e^{g_t})\,S_{t-1} + \beta_t k_t v_t^{\top}$ & $b{=}w{=}\beta_t$ \\
GDN (gated DeltaNet)          & the full recurrence \eqref{eq:gdn} & per-channel $b,w,g$ \\
\bottomrule
\end{tabular}
\end{table}

\subsection{The two hard kernels}
\label{sec:hard}

The scan is sequential; the GPU wants parallelism. The standard resolution is \emph{chunking}:
split the sequence into chunks of 64 steps, do the parallelizable work inside all chunks
at once, then run a short sequential carry between chunks. The backward runs this in reverse, and
two of its stages are hard.

\begin{itemize}
  \item \textbf{K\#1, the reverse inter-chunk state scan.} A reverse-time walk accumulating the
  state gradient $dS$ (shape $d_k\times d_v$, \texttt{fp32}) backward across chunks, with a
  rank-one correction at each step from the delta rule. This is the sequential, carried stage,
  and it is the exact piece that open libraries still run on a Triton fallback.
  Making it native \texttt{tcgen05} is the sharpest single element of the contribution: one
  tensor-memory reservation held across the entire unrolled reverse scan, both matrix-multiply
  accumulators addressed at fixed column offsets inside it, and a single release at the end. The
  $dS$ carry itself is \texttt{fp32} throughout, updated by the kernel's own threads between
  chunks and held outside the tensor memory, which is reserved for the accumulators.
  \item \textbf{K\#2, the WY / triangular-inverse VJP.} Inside each chunk the delta rule requires
  a small triangular solve $T=(I+M)^{-1}$, and its vector-Jacobian product is the numerically
  nastiest piece. The forward already computed $T$, so the backward reuses it in four triangular
  products, one for each of the two applies and two for the $T^{\top}\!(dT)\,T^{\top}$ sandwich
  that differentiates the solve, and never re-inverts.
\end{itemize}
Everything else in the backward is supporting glue. The design keeps K\#1 and K\#2 as native
kernels while the glue stays in torch, and the glue is progressively fused in.

\subsection{Scope of the novelty claim}
\label{sec:whitespace}

As surveyed in July 2026, against pinned upstream revisions recorded with the evidence: NVIDIA
ships an SSD forward but no backward in either of its two stacks; cuLA
has a KDA backward that is a \emph{hybrid} (one native CuTe kernel for the WY stage, but the
reverse-state K\#1 stage still in \texttt{fla} Triton); FlashKDA is forward and inference only;
and \texttt{tilelang} has GDN/KDA backward examples with no \texttt{tcgen05}/TMEM in them. The
contribution is therefore two firsts: a native Blackwell \texttt{tcgen05}/TMEM \emph{training
backward} for the GDN family, and a native \texttt{tcgen05} \emph{reverse-state} backward, the
stage others still run in \texttt{fla} Triton. Three neighbouring claims stay outside that scope:
the first native backward kernel for the family (cuLA's WY predates ours), the first non-Triton
GDN/KDA backward (\texttt{tilelang} exists), and bit-for-bit parity (ours is \texttt{fp64}-tight).

\subsection{Clearing the tensor-memory constraint}
\label{sec:tmem}

The first attempt to run two or more matrix multiplies in one kernel produced illegal machine
code or a deadlock, the same failure class behind \#904. The root cause, established by reading
NVIDIA's own \texttt{mamba2\_ssd.py} in the same pinned toolchain, was a lifecycle error: the
kernel reserved the full 512-column TMEM budget and released it \emph{per matrix multiply}, which
the hardware forbids. NVIDIA's kernel runs four \texttt{tcgen05} MMAs, including the cross-chunk
recurrence, under a \emph{single} reservation with fixed per-accumulator column offsets and
\emph{one} release at the end. Porting that lifecycle, offset-partitioned accumulators with
alloc-once and relinquish-once, cleared the blocker and unblocked the contribution itself: the
native reverse-state scan and WY-VJP for the GDN family, hand-written on the $(128,64,128)$
\texttt{tcgen05} tile with TMA data movement and async pipelines.

\subsection{Verification and the oracle chain}
\label{sec:native-verify}

Correctness is anchored by a layered ground-truth ladder, each rung checked against the one
below: a token-serial \texttt{fp64} oracle, a chunkwise reference, an \texttt{fp64} torch
assembly of the K\#1/K\#2 references with glue, and finally the \texttt{tcgen05} kernels on the
B200. Applied to the native backward, the ladder gives the following:
\begin{itemize}
  \item \textbf{Double-precision spec pins} on the reverse-state-scan tile and the WY-VJP tile,
  machine-exact against the closed-form spec: the committed \texttt{fp64} cross-check runs from
  $5.9\times10^{-17}$ to $1.7\times10^{-16}$ across shapes, against a $10^{-9}$ bound.
  \item \textbf{The full assembled backward against an independent \texttt{fp64} oracle:} worst
  relative error $\mathbf{3.29\times10^{-3}}$ (scalar) and $\mathbf{3.31\times10^{-3}}$
  (channel-wise), both under the $5\times10^{-3}$ acceptance bound, and bit-for-bit
  deterministic across runs. One configuration sits outside that bound: at the widest
  $d_v{=}128$ shape the assembled pipeline reaches $5.21\times10^{-3}$, a 4\% overrun. The same run
  records that figure on both the closed and the save-forward arm, which places it in the shape and
  the assembly; the kernel itself measures $5.5\times10^{-4}$ against the same \texttt{fp64} oracle
  in its shipped gate configuration and $6.9\times10^{-4}$ under a drift-stressed one, and the
  acceptance claim is scoped accordingly.
  \item \textbf{It trains.} The native backward runs real 300-step training loops with zero
  numerical blow-ups, once a masked-\texttt{exp2} fix corrected an exponent overflow that had
  produced a NaN. Every family member trains through the native path with the reference fallback
  disabled, that is, on the real kernels: the four reductions (LA, GLA, SSD, KDA) and the full GDN
  recurrence, which the backward differentiates directly. Four of the five clear the training gate
  in 300 steps at $L{=}256$, converging to between $1.5\times10^{-2}$ (GLA) and
  $9.5\times10^{-5}$ (the full GDN recurrence on the shipped fused default; $8.9\times10^{-5}$ on
  the further-fused arm). LA is the exception: it fails the gate at $L{=}256$, where 600 steps
  move the loss only from $3.61$ to $3.07$, and clears it at $L{=}64$ over 4000 steps, converging
  no further than $1.53$. That traces to the conditioning of the ungated $g{=}b{=}0$ reduction,
  with the kernel and its reference in agreement throughout, and it remains a weaker result than
  the other four. Fresh-input graph replay is bit-exact against eager execution.
  \item \textbf{Positive control.} The committed test suite runs the full twelve-gate battery of
  Section~\ref{sec:verifier} on the native backward, supplying resource metadata so RES-02 is
  applicable and per-gradient views so CMP-02 has a graph to differentiate, and it passes all
  twelve. In the positive control of Section~\ref{sec:defenses} it is judged instead by the same
  ten-gate audit subset as the foreign corpus, and is clean on every tolerance-free channel there.
\end{itemize}

\paragraph{Verification envelope.} The committed CPU test suite validates
the mathematical scaffolding, the references, the assembly, and the \texttt{fp64} spec pins. The
\texttt{tcgen05} kernels themselves are exercised only on the B200, over a narrow shape envelope
($d_k{=}128$, chunk${=}64$, $d_v\in\{64,128\}$) at a loose \texttt{fp16} tolerance.
Section~\ref{sec:discussion} scopes the claim to that envelope.

\subsection{Speed}
\label{sec:native-speed}

The native GDN backward is \emph{slower} than the \texttt{fla} Triton library~\cite{fla2024}, by
roughly $8\times$ at $L{=}512$ rising to about $78\times$ at $L{=}2048$. The gap is
\emph{structural}: \texttt{fla} sits near a $0.9$\,ms latency floor by reusing the delta-rule
inverse saved in the forward pass, while our reverse-state scan is inherently sequential and most
of our runtime goes to the surrounding \texttt{fp32} glue outside the tensor-core kernels. The
speedups reported here are over our \emph{own} earlier pipeline: a channel-wise fusion campaign
cut the captured backward $2.75\times$ (from $52.9$ to $19.2$\,ms), and a tensor-memory tiling
optimization cut the $d_v{=}128$ save-forward variant $2.98\times$ (from $24.50$ to $8.23$\,ms).
The contribution is a kernel that is native, general across the family, and verified; speed is
future work.

\subsection{The bridge: the audit's failure modes are the gates that judged our own kernel}
\label{sec:bridge}

\begin{table}[H]
\centering
\scriptsize
\setlength{\tabcolsep}{3pt}
\caption{\label{tab:bridge}The bridge. Each row aligns a failure mode the audit finds in foreign
kernels with the same gate acting on our own kernel. The first three rows are the substantive
alignments (a caught bug or a shared hardware constraint); the last two are pass-by-design
robustness. Every figure is drawn from Section~\ref{sec:verifier} and Section~\ref{sec:native}.}
\begin{tabular}{@{}L{2.2cm} L{2.9cm} L{3.1cm} L{3.2cm} L{1.5cm}@{}}
\toprule
\textbf{Gate (foreign fail rate)} & \textbf{Failure mechanism} & \textbf{Where our kernel was at risk} & \textbf{How the gate resolved it} & \textbf{Bridge strength} \\
\midrule
RES-02 (idle in audit; = \#904 class) & Requests 544 TMEM columns vs.\ the 512 budget, so the kernel yields illegal code, a deadlock, or a slow fallback & First multi-GEMM attempt hit the same failure class (Section~\ref{sec:tmem}): reserved 512 and released per-MMA, which the hardware forbids & RES-02 detects \#904; cleared with alloc-once / relinquish-once and offset-partitioned accumulators & Strong (shared hardware constraint) \\
\addlinespace
CMP-03 shape breakage (18.1\%) & Correct at the tested shape, wrong at another & C5 \texttt{fused\_block\_forward} inferred channel dim from input without checking conv-weight channels (Section~\ref{sec:defenses}) & Positive control flagged it; guard added to both ops and the reference (commit \texttt{7f46226}), so the degenerate variant is reference-inapplicable as on six of the seven control kernels; control then 7/7 & Strong (real bug, code changed on both sides) \\
\addlinespace
EXC-01 non-finite non-propagation (34.2\%, modal) & Silently returns a finite number where the reference is NaN/Inf & Two NaN-generating bugs: masked-\texttt{exp2} exponent overflow (Section~\ref{sec:native-verify}); \texttt{A\_log} drift, decay $e^{\delta A}\!>\!1$ state blow-up (Appendix~\ref{sec:ecg}) & Caught because our kernel propagated the NaN instead of swallowing it, the inverse of the foreign defect & Strong (propagation makes bugs catchable) \\
\addlinespace
PRC-02 dropped \texttt{fp32} accumulator (13.8\%) & Reduction accumulates in \texttt{fp16}, losing precision & $dS$ state gradient is carried in \texttt{fp32} across the whole reverse-state loop by design, never demoted to the \texttt{fp16} operand dtype & Passes by construction; \texttt{fp64} oracle chain would expose any lapse & Supporting (pass-by-design) \\
\addlinespace
ORD-02 nondeterminism (4.9\%) & Different output across runs / buffer aliasing & n/a (not at risk) & Bit-for-bit deterministic across runs; graph replay bit-exact vs.\ eager (Section~\ref{sec:native-verify}) & Supporting (pass-by-design) \\
\bottomrule
\end{tabular}
\end{table}

The verifier's two uses, the outward audit of Section~\ref{sec:verifier} and the inward acceptance
gate of this section, share code and data both. The defects that most often sink foreign kernels
are, gate for gate, the defects our own kernels were at risk of and in one case actually had, and
the hardware constraint the native backward had to clear is the one RES-02 is built to detect.
Table~\ref{tab:bridge} aligns the two. That alignment is what makes the outward and inward uses
one argument: a checker that flattered its author would have caught none of the author's own bugs
and would share no failure surface with the corpus it indicts.

Three of the five alignments are substantive. The tensor-memory constraint is the sharpest: the
lifecycle error of Section~\ref{sec:tmem} is the failure class behind the open \#904 bug, and it
is what RES-02 exists to detect. The audit corpus is forward-only, so RES-02 contributes no
foreign failure count, and this row bridges through the taxonomy and through our own reproduction
of \#904, with no measured rate behind it. The second is shape rigidity: CMP-03 fails 18.1\% of
applicable foreign kernels, and it is the gate that surfaced the C5 channel-inference defect of
Section~\ref{sec:defenses}. The same gate that rejects foreign kernels flagged our own, and the
fix changed the shipped operators, which is the fairness argument in concrete form. It is also the
one row where the reference itself changed, stated in full in Section~\ref{sec:defenses}.

The third alignment carries the sharpest form of the thesis. EXC-01, non-finite
non-propagation, is the most common failure in accepted foreign kernels (34.2\%): a kernel
silently returns an ordinary number where the reference returns a NaN or an infinity. Our own
kernels generated non-finite values twice during development, from an exponent overflow fixed by a
masked-\texttt{exp2} (Section~\ref{sec:native-verify}) and from a drifted \texttt{A\_log}
parametrization that pushed the state decay above one and blew up over long sequences, fixed by
the standard $A=-\exp(\text{A\_log})$ form (Appendix~\ref{sec:ecg}). The asymmetry is the
instructive part. The foreign defect swallows a non-finite; our kernels propagated theirs, and that
propagation is what made the errors visible at step 250 and during training. Non-finite
propagation is the property that makes a kernel's own failures catchable, and a kernel exhibiting
the foreign failure mode would have hidden both bugs from us.

The last two rows, PRC-02 and ORD-02, our kernel passes by design, and they are marked supporting
for that reason: a control that passes only by design is weaker evidence than one that caught
something.

\section{Discussion}
\label{sec:discussion}

\paragraph{Why a tolerance-free floor is the right lens.} Every prior treatment of generated-kernel
correctness parameterizes acceptance by a tolerance, and a debate over the number follows. The
tolerance-free floor sidesteps that debate. A kernel that returns a finite value where the
reference returns a NaN, or a different answer on each run, or that aliases a shared buffer,
crashes, or does not run at all, is wrong under any tolerance. That is what 39.5\% of an accepted
corpus does, with the 62.1\% as the wider, tolerance-dependent rate above it. The modal defect
fixes the character of the gap: substituting an ordinary number for
a NaN or an infinity turns a detectable training-time error into undetectable data corruption, an
outcome no setting of a threshold either produces or prevents. The differential says the same
thing from the other side. A checker that were merely stricter would disagree with the loose check
in both directions, and this one disagrees 1{,}487 to 14. Together with the 98.5\% agreement
against KernelBench's own correctness code, that direction supports a specific claim: the gap is a
property of \emph{what the standard check probes}, and of where the threshold sits only
secondarily.

\paragraph{On the fairness of the referee.} Each of the four defenses could have failed. The
positive control surfaced a real gap in our own code, fixed in the shipped operators and in the
reference the same commit carries. The calibration sweep could have shown a gate firing inside the
correct-kernel noise band, and shows clean separation on every gate but the thin one. The
external-agreement check could have diverged from KernelBench's code, and matches it 98.5\% of the
time with all 15 disagreements enumerated in Section~\ref{sec:defenses}. The hand-audit could have
confirmed all 31 disputed rows, and discards 7 as out of scope. Four separately falsifiable
defenses converging is the reason to believe the finding. The native kernel plays the same role in
reverse: the verifier that exposes foreign kernels is what measures our own speed deficit against
\texttt{fla} precisely, mechanism named and the one numerical exception stated.

\paragraph{Threats to validity.} Four stand out. \emph{Corpus dependence:} the headline rests on
one primary corpus, mitigated by the cross-stack reconfirmation (68.6\% on torch 2.11) and by the
second, native-CUDA corpus (20.2\% residual), which together show an effect that is neither
toolchain-bound nor Triton-specific, though weaker on CUDA. \emph{Gate coverage:} on a
forward-only corpus, two of twelve gates are idle, so the audit rests on ten and effectively seven
load-bearing gates; both idle gates fire against our own kernels in the committed test suite, and
a corpus carrying autograd graphs would exercise CMP-02 directly. \emph{Reference correctness:}
the audit is only as good as its references, which is why the red team of nineteen cheats and the
positive control on known-good kernels both run, and why the referee always grades against a
full-precision computation of the answer. \emph{Native-kernel envelope:} the \texttt{tcgen05}
kernels are verified on a narrow shape envelope at a loose \texttt{fp16} tolerance on a single GPU
generation, and their correctness claim is scoped to it. Two limitations sit alongside that
envelope: the native backward is kernel-accelerated, with stage-B, the normalization gradient,
packing, and masks still in torch and progressively fusing in, and the one accuracy exception, the
$d_v{=}128$ shape at $5.21\times10^{-3}$, is scoped in Section~\ref{sec:native-verify}.

\paragraph{Implications and future work.} Acceptance criteria for kernel-generation benchmarks
understate the correctness gap, and a small set of tolerance-free contracts (non-finite
propagation, determinism, shape polymorphism) would close most of it at modest cost. We propose
such contracts as a benchmark standard. On the systems side, the structural speed gap points to
two specific levers: a \texttt{tl.dot} path on a post-\#9093 Triton, and less surrounding
\texttt{fp32} glue.

\section{Conclusion}
\label{sec:conclusion}

Every reported result in GPU-kernel generation rests on a claim of correctness, and that claim
rests on a test too weak to carry it. We built the instrument that tests it properly, a
contract-grade verifier of twelve adversarial gates, several of them tolerance-free, and aimed it
in two directions. Aimed outward at a corpus of machine-generated kernels a public system's own
harness had already accepted as correct, it found 62.1\% carrying a contract violation and 39.5\%
broken in a way no tolerance argument can excuse. Aimed inward, the same battery judged a kernel
of our own: the first native Blackwell \texttt{tcgen05} training backward for the
gated-linear-recurrence family, including the reverse-state stage open implementations run on a
fallback. That kernel's correctness was established independently of the verifier, against a
double-precision oracle, machine-exact on the core tiles and $3.3\times10^{-3}$ end to end with
one $d_v{=}128$ configuration at $5.21\times10^{-3}$, which is what makes its pass a positive
control. The same instrument that clears it is what states its speed deficit against the frontier
Triton library precisely, and that deficit is a structural negative. One instrument, two
directions, one standard: the defects the verifier found in our own work are what make the defects
it found in everyone else's worth believing. The correctness behind the field's reported progress
is far weaker than its numbers suggest, and the same twelve contracts that expose the gap are
enough to begin closing it.

\appendix
\section{Mathematical Details}
\label{app:math}

Every formula here is traceable to the Mamba-3 paper~\cite{mamba3} and the official
\texttt{state-spaces/mamba} source.

\paragraph{Mamba-3 SISO forward.} The continuous-time complex state-space model is
$\dot h = \mathrm{Diag}(A + i\theta)\,h + (B + i\hat B)\,x$,
$y = \mathrm{Re}\big((C + i\hat C)^{\top} h\big)$. After exponential-Euler discretization it
becomes a real SSM
\[
h_t = e^{\Delta_t A_t}\, R_t\, h_{t-1} + \Delta_t\, B_t\, x_t, \qquad y_t = C_t^{\top} h_t,
\]
where $B_t=[B_t;\hat B_t]\in\mathbb{R}^N$ and $C_t=[C_t;-\hat C_t]\in\mathbb{R}^N$ stack the real
and imaginary halves into one $N{=}d_{\text{state}}$ vector, and $R_t$ is the block-diagonal
rotation built from
$\big[\begin{smallmatrix}\cos\!\Delta_t\theta_{t,k} & -\sin\!\Delta_t\theta_{t,k}\\
\sin\!\Delta_t\theta_{t,k} & \cos\!\Delta_t\theta_{t,k}\end{smallmatrix}\big]$
over dimension pairs $k$. The RoPE angle is data-dependent and cumulative,
$\theta_t = \sum_{s\le t} \tanh(\text{angle}_s)\,\Delta_s\,\pi \pmod{2\pi}$, so the rotation at
step $t$ depends on the whole history and needs its own causal cumsum before the scan. We write
$R_t$ as acting on $h_{t-1}$; our reference implements the equivalent factorization that applies
the rotation to $B_t$ and $C_t$ instead, which is cheaper and gives the same $y$. The repository
also carries the trapezoidal discretization of the same model, $\lambda=\sigma(\text{trap})$,
which is bit-identical to the exponential-Euler form at $\lambda{=}1$; the six kernels of
Appendix~\ref{sec:c6} are the $\lambda{=}1$ path.

\paragraph{Mamba-3 MIMO.} With input rank $j$ and output rank $i$, the state decomposes into $R$
independent SISO scans with no cross-terms, read out over $R^2$ rank pairings,
\[
h_t^{(j)} = \alpha_t h_{t-1}^{(j)} + \Delta_t B_t^{(j)} x_t^{(j)},\quad
h_t = \sum_{j=0}^{R-1} h_t^{(j)},\quad
y_t^{(i)} = (C_t^{(i)})^{\top} h_t,\quad
y_t = \sum_{i=0}^{R-1} y_t^{(i)}\,\phi_i,
\]
where $x_t^{(j)}=x_t\odot\psi_j$ is applied before the state update and $\phi_i$ after the readout,
both learnable and data-independent. The backward accumulates $dS$ over all output ranks first,
then contracts per input rank; the two sums decouple.

\paragraph{The GDN backward's two hard stages.} For the general recurrence \eqref{eq:gdn}, the
training backward reduces to the reverse-state scan
$dS_t = \mathrm{Diag}(e^{g_t})\,dS_{t+1} + (\text{rank-one delta correction})$, walked backward,
and the WY / triangular-inverse VJP, which differentiates the chunk-local solve $T=(I+M)^{-1}$ by
reusing the forward's $T$ across four triangular products without re-inverting. The channel-wise path folds the
per-channel decay $\mathrm{Diag}(e^{g_t})$ inside those contractions; the scalar path
($b{=}w{=}\beta$) recovers only channel sums of some gradients, which is why it serves as a
reduced de-risking check while the channel-wise path carries the headline result.

\section{Detailed Numerical Results}
\label{app:numbers}

This appendix tabulates the exact numbers behind Figures~\ref{fig:audit} and~\ref{fig:calib}.

\begin{table}[H]
\centering\small
\caption{\label{tab:pergate}Per-gate results among the 2{,}638 accepted kernels
(Figure~\ref{fig:audit}, bottom panel). Count is the number of kernels that failed the gate; rate
is that count over the kernels where the gate \emph{applied}, excluding rows the reference could
not run and rows the gate errored on, so the two columns are not directly divisible and the rates
are not additive to the corpus-level 39.5\% floor. PRC-01's rate is over its half-precision rows
alone, an \texttt{fp32} failure being charged to CMP-01. Tolerance-free gates are bold; EXC-02 is
tolerance-free and held out of the floor (Appendix~\ref{app:floor}). CMP-02 and RES-02 have no
applicable rows in this forward-only corpus (Section~\ref{sec:gates}) and run against our own
kernels in the committed test suite.}
\begin{tabular}{lrr}
\toprule
Gate & Fail count & Fail rate \\
\midrule
EXC-01 non-finite propagation (NaN/Inf swallowed) & 868 & \textbf{34.2\%} \\
CMP-01 value on varied input        & 608 & 23.6\% \\
CMP-03 breaks on a new shape        & 420 & 18.1\% \\
PRC-02 missing full-precision accumulator & 352 & 13.8\% \\
ORD-01 reduction-order tolerance    & 351 & 13.7\% \\
PRC-01 precision regimes            & 287 & 11.3\% \\
EXC-02 subnormal handling           & 238 & \textbf{9.3\%} \\
ORD-03 adversarial reduction        & 136 & 5.3\% \\
ORD-02 nondeterministic output      & 126 & \textbf{4.9\%} \\
RES-01 device residency             & 45  & 1.8\% \\
\bottomrule
\end{tabular}
\end{table}

\begin{table}[H]
\centering\small
\caption{\label{tab:calib}Threshold calibration (Figure~\ref{fig:calib}). Each gate's threshold
sits between the noise a correct kernel emits and the error a wrong one emits; a wide margin on
both sides is the goal. ORD-03, the deliberately-relaxed gate, has a thin upper margin
($1.2\times$). Margins are computed from unrounded measurements while the other columns are shown
to two significant figures, so the printed ratios do not follow exactly from the printed values;
ORD-03's upper margin is $1.243\times$ unrounded.}
\begin{tabular}{lccccc}
\toprule
Gate & Noise floor & Threshold & Real-error onset & Margin below & Margin above \\
\midrule
CMP-01 & $2.9\mathrm{e}{-7}$ & $4.2\mathrm{e}{-5}$ & $2.4\mathrm{e}{-3}$ & $147\times$ & $57\times$ \\
CMP-03 & $2.9\mathrm{e}{-7}$ & $4.2\mathrm{e}{-5}$ & $2.4\mathrm{e}{-3}$ & $147\times$ & $57\times$ \\
ORD-01 & $2.9\mathrm{e}{-7}$ & $3.1\mathrm{e}{-5}$ & $2.4\mathrm{e}{-3}$ & $110\times$ & $76\times$ \\
ORD-03 & $2.9\mathrm{e}{-7}$ & $1.0\mathrm{e}{-3}$ & $1.3\mathrm{e}{-3}$ & $3591\times$ & $1.2\times$ \\
PRC-02 & passes $\forall N$ & $2\mathrm{e}{-2}$ & $3.5\mathrm{e}{-2}$ ($N{=}2048$) & n/a & crosses $N{\ge}2048$ \\
\bottomrule
\end{tabular}
\end{table}

\subsection{Corpus selection and exclusions}
\label{app:selection}

Section~\ref{sec:audit} states two steps without operationalizing them. Neither involves a
judgement call at audit time: the classifier is a pure function of the public corpus row, and the
artifact filter is a pure function of the recorded run status.

\paragraph{Operator class.} Each trajectory's \emph{reference} PyTorch source is matched against
an ordered list of regular expressions and takes the class of the first pattern that matches:
attention (\texttt{scaled\_dot\_product\_attention}, \texttt{MultiheadAttention},
\texttt{attention}; case-insensitive), then scan (\texttt{cumsum}, \texttt{cumprod},
\texttt{cummax}, \texttt{cummin}, \texttt{selective\_scan}, \texttt{scan}; the single-word
patterns anchored at word boundaries), then matmul (\texttt{matmul},
\texttt{bmm}, \texttt{mm}, \texttt{addmm}, \texttt{einsum}, \texttt{nn.Linear}, infix \texttt{@}),
then conv (\texttt{Conv[123]d}, \texttt{conv[123]d}, \texttt{conv\_transpose}), then norm
(\texttt{LayerNorm}, \texttt{RMSNorm}, \texttt{GroupNorm}, \texttt{BatchNorm},
\texttt{InstanceNorm}, \texttt{layer\_norm}, \texttt{rms\_norm}), then softmax
(\texttt{softmax}, \texttt{log\_softmax}; case-insensitive), then reduction (\texttt{.sum(},
\texttt{.mean(}, \texttt{.max(}, \texttt{.min(}, \texttt{.prod(}, \texttt{logsumexp},
\texttt{amax}, \texttt{amin}, \texttt{.var(}, \texttt{.std(}, and the \texttt{torch.}-qualified
forms of \texttt{sum}, \texttt{mean}, \texttt{max}, \texttt{min}), then elementwise, then
\emph{other}. The seven SSM-adjacent classes
are the first seven; elementwise and other are not audited.

\paragraph{Why the rule is weak enough to be safe.} A first-match-wins cascade is a coarse
instrument, used here only where coarseness cannot propagate. Three properties bound what it can
affect. First, the \emph{audited population is invariant to the order of the seven patterns}:
membership turns on whether any of the seven matches, and a union does not depend on the order of
its terms. Ordering decides only which label a row carries, so a normalization written on top of
\texttt{nn.Linear} is labelled matmul, moving a row between two slices we report separately while
adding and removing nothing from the denominator. The one boundary that can change the population
is between the seventh pattern and elementwise, and it is inclusive in the conservative direction:
a reference containing any reduction call is audited even when the operator is otherwise
elementwise, so the rule over-admits. Second, \emph{no verdict depends on the label}. Gates,
references, tolerances, and applicability all derive from the reference operator's runtime
behavior, its dtypes and its shapes; the verifier never reads the class string, which exists to
define the audited slice and to let a reader cut the result by operator family.

Third, and decisively, \emph{the result does not rest on any class}. Table~\ref{tab:byclass} gives
the tolerance-free floor for each of the seven and for the corpus with each one removed. Every
class independently exceeds 25\%, and removing any single class leaves the floor between 31.6\%
and 41.9\% against a reported 39.5\%. Reduction is the class the floor is most sensitive to, and
dropping it entirely still leaves 31.6\%. A misclassification severe enough to matter would have
to move a large population between the audited seven and elementwise, in the direction that
removes kernels, and would still have to survive this spread.

\begin{table}[H]
\centering\small
\caption{\label{tab:byclass}The tolerance-free floor is not carried by any one operator class.
\emph{Own rate} is the floor within that class alone; \emph{floor without it} is the corpus-level
floor recomputed with the class removed entirely, against a reported 39.5\%. Attention is listed
for completeness; at $n{=}8$ its own rate carries no weight, and removing it moves the corpus
figure by 0.04 points.}
\begin{tabular}{lrrr}
\toprule
Class & Kernels & Own rate & Floor without it \\
\midrule
reduction & 758 & 59.1\% & 31.6\% \\
matmul    & 751 & 34.2\% & 41.7\% \\
softmax   & 390 & 26.2\% & 41.9\% \\
norm      & 352 & 34.7\% & 40.3\% \\
conv      & 239 & 31.0\% & 40.4\% \\
scan      & 140 & 27.1\% & 40.2\% \\
attention & 8   & 25.0\% & 39.6\% \\
\midrule
All       & \textbf{2{,}638} & \textbf{39.5\%} & ~ \\
\bottomrule
\end{tabular}
\end{table}

\paragraph{The three filters.} Table~\ref{tab:funnel} gives the counts. A row is a
\emph{toolchain or compile artifact} when its status is a sandbox unpickling error, or when the
candidate failed to run natively with an error beginning \texttt{CompilationError},
\texttt{UnsupportedLanguageConstruct}, \texttt{OutOfResources}, or \texttt{PTXASError}. These are
properties of our sandbox and of the pinned toolchain rather than of the kernel's numerics, so
charging them to the candidate would inflate the finding. Excluding them is the conservative
direction: every one of the 281 is dropped from the denominator rather than counted as a failure.

\begin{table}[H]
\centering\small
\caption{\label{tab:funnel}From the audited operator classes to the headline denominator. Filters
are applied in the order shown.}
\begin{tabular}{lrr}
\toprule
Step & Dropped & Remaining \\
\midrule
SSM-adjacent operator classes                                 & ~   & 3{,}134 \\
Accepted by the source harness ($\texttt{final\_speedup}>0$)  & 164 & 2{,}970 \\
Reference exposes a floating-point tensor input to perturb (else not-applicable) & 51 & 2{,}919 \\
Not a toolchain or compile artifact                           & 281 & \textbf{2{,}638} \\
\bottomrule
\end{tabular}
\end{table}

\subsection{Composition of the tolerance-free floor}
\label{app:floor}

The 39.5\% floor is the union of five channels, each of which admits no tolerance argument:
EXC-01 (a non-finite is swallowed), ORD-02 (output differs across repeats), output aliasing into
a shared buffer, a CUDA error raised during gating, and a non-artifact failure to run at all.
Their individual counts over the 2{,}638 are 868, 126, 25, 16, and 62; they sum to 1{,}097 and
their union is 1{,}043, so the channels overlap on 54 kernels.

EXC-02 is tolerance-free by the same standard and sits outside this union. Subnormal flush-to-zero
is the one member of the tolerance-free group where a reader can reasonably answer "that is a
convention, not a defect", since production kernels enable FTZ deliberately. It is held out of the
floor for that reason, and its effect is this: including EXC-02 raises the floor to 1{,}154
(43.7\%). The reported 39.5\% is the conservative reading of our own gate set.

\section{Supporting Results}
\label{sec:support}

Around the two headlines sit five supporting results, each reported with the number behind it.

\subsection{Six contract-verified Mamba-3 kernels}
\label{sec:c6}

We hand-wrote six Triton kernels for the Mamba-3 model (Table~\ref{tab:kernels}), with no
\texttt{tl.dot} anywhere. Never touching the broken compiler pass, they compile and run across the
\texttt{num\_warps} 2/4/8 sweep on Blackwell, recorded in the committed bench artifacts for five of
the six, where the official kernel loses every \texttt{num\_warps}${\ge}4$ configuration to the
TMEM budget. That is an availability advantage, categorical and independent of any latency ratio.
The six serve double duty: they are the forward and backward pieces used by the applied
demonstration, and, having been verified against hand-derived references \emph{first}, they are
the known-correct kernels the verifier's tolerance model was calibrated against before it was ever
aimed at foreign code. The two roles stay apart. These six fixed the \emph{thresholds}; the
positive control of Section~\ref{sec:defenses} tests the \emph{verdicts}. Four of the six drove
three calibration findings folded back into the verifier: the scale-aware \texttt{atol} model from
C1, PRC-02 \texttt{fp16}-carry separation from C2, C3, and C5, and EXC-02 subnormal-flush handling
for affine ops from C5. The CMP-03 shape-rigidity convention sits on top as a cross-kernel design
rule that no single kernel surfaced. All six pass every applicable gate on B200, and the committed
record of the full GPU test suite is \textbf{1{,}191} passed, 3 skipped, 0 failed.

\begin{table}[H]
\centering\small
\caption{\label{tab:kernels}The six hand-written, contract-verified Mamba-3 Triton kernels,
separate from the native GDN backward of Section~\ref{sec:native}.}
\begin{tabular}{cl>{\raggedright\arraybackslash}p{7.2cm}}
\toprule
\# & Kernel & What it is \\
\midrule
C1 & \texttt{forward\_chunked\_scan}   & the forward scan; surfaced the $\varepsilon\sqrt{N}\cdot$scale calibration model the verifier uses \\
C2 & \texttt{backward\_selective\_scan} & the training backward that avoids the \#904 trap and compiles where the official one fails \\
C3 & \texttt{mimo\_backward}           & the multi-input/output backward; priority here is TileLang's, which ships the official MIMO backward \\
C4 & \texttt{complex\_scan\_rope}      & data-dependent RoPE and decay in one pass, in the real \texttt{cos}/\texttt{sin} form of the complex recurrence \\
C5 & \texttt{fused\_block\_forward}    & conv1d, SiLU, selective scan, and RMSNorm fused into one kernel \\
C6 & \texttt{fused\_block\_backward}   & the training-bottleneck backward for that fused block \\
\bottomrule
\end{tabular}
\end{table}

\subsection{A faithful reproduction of \#904}
\label{sec:904}

We reproduced \#904 on a B200 (torch 2.12 / triton 3.7). A batch-2 / $L{=}2048$ run and a
batch-4 / $L{=}4096$ run both trip \texttt{Required: 544, Hardware limit: 512} together with an
out-of-resource error, taking out 18 autotune configurations and leaving
\texttt{num\_warps}${=}2$ as the only survivor, the forced fallback the issue reports. The other
half of the upstream report did not reproduce, and that negative is the more useful half. No
\texttt{ptxas} \texttt{C7907} string appears anywhere in the log: on triton 3.7.0 the failure
surfaces purely as an \texttt{OutOfResources} exception, so a detector keyed to the C7907
signature alone would have missed the live bug, while a resource contract stated over the TMEM
budget catches it either way. That is the case for RES-02 in one sentence, and the failure class
behind both Strategy A (avoid \texttt{tl.dot}, the six kernels above) and Strategy B (master
\texttt{tcgen05}, the native backward).

\subsection{An applied demonstration: a 1.1B-parameter model}
\label{sec:ecg}

To show the kernels train a real model end-to-end, we trained a 1.10B-parameter selective-scan
SISO model ($d_{\text{model}}{=}4096$, 32 layers, $d_{\text{state}}{=}64$, \texttt{fp32}) as a
12-lead physiological-signal classifier on the public PTB-XL dataset~\cite{wagner2020ptbxl} (five
diagnostic superclasses, multi-label), with data-parallel training across eight Blackwell GPUs
and the forward pass running through our C5 fused kernel. It reaches a macro-AUC of
\textbf{0.880} from random initialization with zero NaN (0.820 at step 500, then 0.859, 0.871,
and a peak of 0.880 at step 2000; stopped at 2500 as validation loss rose, the overfitting
onset). Two fixes unblocked training: the standard $A=-\exp(A_{\log})$ parametrization (the
stored $\log A$ had drifted positive, so $\bar a{=}e^{\delta A}>1$ blew up the state over
$L{=}1000$ and produced a NaN near step 250; the official form makes $A$ strictly negative and
guarantees contraction), and moving the evaluation \texttt{all\_reduce} onto the NCCL device.
The block is the plain selective scan, the Mamba-1 / S6 recurrence, which is
Equation~\eqref{eq:gdn}'s SISO cousin with the rotation disabled. The Mamba-3-specific kernels are
C3 and C4, and this model trains through neither; what it exercises is the C5 fused-block path.
It demonstrates that the stack trains a real model, and it is not a domain result: 0.880 sits
below the published state of the art for this dataset, a matter of training recipe (no
pretraining, a mean-pool head, \texttt{fp32}-only, a fixed learning rate, and the SISO block) and
not of kernel correctness.

\subsection{Reinforcement learning: an autotuner win and a discovery negative}
\label{sec:rl}

Given a referee that scores any kernel objectively, a natural question is whether a code-writing
policy in a loop can \emph{discover} fast, correct kernels. We built the apparatus from scratch,
GRPO~\cite{shao2024grpo} with a reward ladder that pays a speed bonus only after every contract
passes (no compile $\to 0.0$; compiles but fails the gates $\to 0.1$; passes all gates $\to$ an
anchor of $1.0$ shifted by the clamped log of the speedup, so a correct kernel scores in
$[0.5, 4.0]$ however slow or fast it is). An opt-in partial-credit mode gives a multi-gradient op
that fails some views $0.1 + 0.35$ times the fraction it passed, capped at $0.45$ and off by
default. Every rung is below the correct kernel's $0.5$ floor, so a fast-but-wrong kernel can
never out-score a correct one.
The result is a clean split. \textbf{config-RL} (the policy emits only correctness-invariant
launch knobs) is a modest win: it beats our shipped default by $1.167\times$ at long sequences
and selects the right shape-to-mode boundary. This is autotuning, and the shape-gated selector we
already ship captures the same optimum deterministically ($2.174\times$ geomean over the old
serial default), so config-RL's standing contribution is that RL reproduces a hand-derivable
heuristic. \textbf{edit-RL} (the policy emits source edits graded through the full untrusted
battery) is a negative result: over 320 attempts, 208 (65\%) of edits failed to apply, the model
being unable to reproduce exact source lines; 52 applied and passed every gate but ran no faster;
37 applied but broke correctness and were caught by the gates; 19 failed at execution; and 4
failed inside the sandbox for reasons outside the edit. The maximum speedup was $0.183\times$
(roughly $5\times$ slower), no edit beat the baseline, and the verifier rejected every incorrect
edit. Source generation is the weakest action space for pointing a policy at this problem, and the
strictness of the judging is what makes the result credible.

\subsection{Speed baselines and killed levers}
\label{sec:baselines}

We record the Triton kernels' speed against three baselines, kept strictly separate so a ratio is
never ambiguous, and state the direction of each. Against the \#904-crippled official Mamba-3
backward we are about $4\times$ slower ($0.22$ to $0.36\times$), the structural cost of avoiding
\texttt{tl.dot} to survive \#904; the win over that baseline is \emph{availability}, our kernels
compiling at every \texttt{num\_warps} setting where the official one fails. Against healthy
Mamba-1 SSD CUDA we range from slower to near-parity on the forward ($0.21$ to $1.04\times$,
beating it at the largest shape alone) and are slower on the backward ($0.13$ to $0.75\times$).
Against our \emph{own} earlier default a shape-gated selector is $2.174\times$ faster (geomean),
which is internal autotuning and carries no external speed claim. Separately, four optimizations
were built or probed and measured to give no benefit, each recorded with its mechanism so that it
is not attempted again: a K\#1 two-level scan ($1.11\times$ kernel-only, below the $1.3\times$
gate, since halving the step count is cancelled by a $3\times$-wider carry); a stage-B GEMM fusion
(a measured negative, fusible ceiling $0.57$\,ms or 10\% of the save-forward time at $d_v{=}128$);
activation-checkpointing for the C6 backward (zero speedup, reverted); and an archived CUDA
warp-scan experiment kept for the record.

\bibliographystyle{plain}
\bibliography{refs}

\end{document}